\documentclass[11pt,a4paper]{article}
\usepackage[hyperref]{AACL-IJCNLP2020}
\usepackage{times}
\usepackage{latexsym}

\usepackage{microtype}

\aclfinalcopy 

\usepackage{soul}
\usepackage{url}
\usepackage{graphicx}
\usepackage{amsmath}
\usepackage{amsthm}
\usepackage{booktabs}
\usepackage{amssymb}
\def\argmax{\mathop{\rm argmax}}%
\def\argmin{\mathop{\rm argmin}}%
\usepackage{amsmath} 
\usepackage{graphicx}  
\usepackage{subfigure}
\usepackage{multirow}
\usepackage{bbm}
\usepackage[noend]{algpseudocode}
\usepackage{CJKutf8}
\usepackage{pifont}
\usepackage{tabu}
\usepackage{array}
\usepackage{adjustbox}

\title{A General Framework for Adaptation of Neural Machine Translation to Simultaneous Translation }

\author{
 Yun Chen\thanks{\;\;Part of the work was done when Yun is working at Huawei Noah's Ark Lab.}\;$^{1}$,
 Liangyou Li$^2$, Xin Jiang$^2$, Xiao Chen$^2$, Qun Liu$^2$\\
 $^1$Shanghai University of Finance and Economics, Shanghai, China\\
 $^2$Huawei Noah's Ark Lab, Hong Kong, China \\
 yunchen@sufe.edu.cn, \{liliangyou, jiang.xin, chen.xiao2, qun.liu\}@huawei.com\\
}

\date{}

\begin{document}
\begin{CJK*}{UTF8}{gbsn}
\maketitle

\begin{abstract}
Despite the success of neural machine translation (NMT), simultaneous neural machine translation (SNMT), the task of translating in real time before a full sentence has been observed, remains challenging due to the syntactic structure difference and simultaneity requirements. In this paper, we propose a general framework for adapting neural machine translation to translate simultaneously. Our framework contains two parts: prefix translation that utilizes a consecutive NMT model to translate source prefixes and a stopping criterion that determines when to stop the prefix translation. Experiments on three translation corpora and two language pairs show the efficacy of the proposed framework on balancing the quality and latency in adapting NMT to perform simultaneous translation. 
\end{abstract}

\section{Introduction}
Simultaneous translation \cite{Fgen2007SimultaneousTO,Oda2014OptimizingSS,Grissom2014DontUT,niehues2016dynamic,Cho2016CanNM,Gu2017LearningTT,Ma2018STACLST}, the task of producing a partial translation of a sentence before the whole input
sentence ends, is useful in many scenarios including outbound tourism, international summit and multilateral negotiations. Different from the consecutive translation in which translation quality alone matters, simultaneous translation trades off between translation quality and latency. The syntactic structure difference between the source and target language makes simultaneous translation more challenging. For example, when translating from a verb-final (SOV) language (e.g., Japanese) to a verb-media (SVO) language (e.g., English), the verb appears much later in the source sequence than in the target language. Some premature translations can lead to significant loss in quality \cite{Ma2018STACLST}. 

Recently, a number of researchers have endeavored to explore methods for simultaneous translation in the context of NMT~\cite{Bahdanau2014NeuralMT,Vaswani2017AttentionIA}. Some of them propose sophisticated training frameworks explicitly designed for simultaneous translation~\cite{Ma2018STACLST,Arivazhagan2019MonotonicIL}. These approaches are either memory inefficient during training~\cite{Ma2018STACLST} or with hyper-parameters hard to tune~\cite{Arivazhagan2019MonotonicIL}. Others utilize a full-sentence base model to perform simultaneous translation by modifications to the encoder and the decoding process. To match the incremental source context, they replace the bidirectional encoder with a left-to-right encoder ~\cite{Cho2016CanNM,Satija2016,Gu2017LearningTT,Alinejad2018PredictionIS} or recompute the encoder hidden states~\cite{zheng-etal-2019-simpler}. On top of that, heuristic algorithms ~\cite{Cho2016CanNM,Dalvi2018IncrementalDA} or a READ/WRITE model trained with reinforcement learning~\cite{Satija2016,Gu2017LearningTT,Alinejad2018PredictionIS} or supervised learning~\cite{zheng-etal-2019-simpler} are used to decide, at every step, whether to wait for the next source token or output a target token. However, these models either cannot directly use a pretrained consecutive neural machine translation (CNMT) model with bidirectional encoder as the base model or work in a sub-optimal way in the decoding stage. 

In this paper, we study the problem of adapting neural machine translation to translate simultaneously. We formulate simultaneous translation as two nested loops: an outer loop that updates input buffer with newly observed source tokens and an inner loop that translates source tokens in the buffer updated at each outer step. For the outer loop, the input buffer can be updated by an ASR system with an arbitrary update schedule. For the inner loop, we translate using the pretrained CNMT model and stop translation with a stopping controller. Such formulation is different from previous work~\cite{Satija2016,Gu2017LearningTT,Alinejad2018PredictionIS,zheng-etal-2019-simpler} which define simultaneous translation as sequentially making interleaved READ or WRITE decisions. We argue that our formulation is better than the previous one in two aspects: (\romannumeral1) Our formulation can better utilize the available source tokens. Under previous formulation, the number of source tokens observed by the CNMT model is determined by the number of READ actions that has been produced by the policy network. It is likely that the CNMT model does not observe all the available source tokens produced by the ASR system. In contrast, the CNMT model observes all the available source tokens when performing inner loop translation in our framework. (\romannumeral2) Previous formulation makes $T_{\eta}$+$T_{\tau}$ READ or WRITE decisions regardless of the ASR update schedule, where $T_{\eta}$ and $T_{\tau}$ are source sentence and translation length, respectively. For an ASR system that outputs multiple tokens at a time, this is computational costly. Consider an extreme case where the ASR system outputs a full source sentence at a time. Previous work translates with a sequence of $T_{\eta}$+$T_{\tau}$ actions, while we translate with a sequence of $T_{\tau}$ decisions ($T_{\tau}-1$ CONTINUE and 1 STOP). 

Under our proposed framework, we present two schedules for simultaneous translation: one stops the inner loop translation with heuristic and one with a stopping controller learned in a reinforcement learning framework to balance translation quality and latency. We evaluate our method on IWSLT16 German-English (DE-EN) translation in both directions, WMT15 English-German (EN-DE) translation in both directions, and NIST Chinese-to-English (ZH$\rightarrow$EN) translation. The results show our method with reinforced stopping controller consistently improves over the de-facto baselines, and achieves low latency and reasonable BLEU scores.
\section{Background}
Given a set of source--target sentence pairs $\left\langle \mathbf{x}_m,\mathbf{y}^*_m\right \rangle_{m=1}^M$, a consecutive NMT model can be trained by maximizing the log-likelihood of the target sentence from its entire source side context:
\begin{eqnarray}
\label{eq:mle}
\hat{\phi} = \argmax_{\phi} \bigg\{  \sum_{m=1}^M \log p(\mathbf{y}^*_m|\mathbf{x}_m; \phi) \bigg\},
\end{eqnarray}
where $\phi$ is a set of model parameters. At inference time, the NMT model first encodes a source language sentence $\mathbf{x}=\{x_1,...,x_{T_\eta}\}$ with its encoder and passes the encoded representations $\mathbf{h}=\{h_1,...,h_{T_\eta}\}$ to a greedy decoder. Then the greedy decoder generates a translated sentence in target language by sequentially choosing the most likely token at each step $t$:
\begin{eqnarray}
\label{eqn:nt}
y_{t}=&\argmax_{y} p(y|y_{<t},\mathbf{x}). 
\end{eqnarray}
The distribution of next target word is defined as:
\begin{eqnarray}
\label{eqn:pb}
&p(y|y_{<t},\mathbf{x})\propto \exp\left[\phi_{\textsc{out}}\left(z_{t}\right)\right] \nonumber \\
&z_{t}=\phi_{\textsc{dec}}\left(y_{t-1}, z_{<t}, \mathbf{h}\right),
\end{eqnarray}
where $z_{t}$ is the decoder hidden state at position $t$. In consecutive NMT, once obtained, the encoder hidden states $\mathbf{h}$ and the decoder hidden state $z_t$ are not updated anymore and will be reused during the entire decoding process.

\section{Simultaneous NMT}
In SNMT, we receive streaming input tokens, and learn to translate them in real-time. We formulate simultaneous translation as two nested loops: the outer loop that updates an input buffer with newly observed source tokens and the inner loop that translates source tokens in the buffer updated at each outer step. 

More precisely, suppose at the end of outer step $s-1$, the input buffer is $\mathbf{x}^{s-1} = \{x_1, ..., x_{\eta \left[ s-1\right]}\}$, and the output buffer is $\mathbf{y}^{s-1} = \{y_1, ..., y_{\tau \left[ s-1\right]}\}$. Then at outer step $s$, the system translates with the following steps: 
\begin{itemize}
    \item[1] The system observes $c_s > 0$ new source tokens and updates the input buffer to be $\mathbf{x}^{s} = \{x_1, ..., x_{\eta \left[ s\right]}\}$ where $\eta \left[ s\right]=\eta \left[ s-1\right]+c_s$.
    \item[2] Then, the system starts inner loop translation and writes $w_s>=0$ target tokens to the output buffer. The output buffer is updated to be $\mathbf{y}^{s} = \{y_1, ..., y_{\tau \left[ s\right]}\}$ where $\tau \left[ s\right]=\tau \left[ s-1\right]+w_s$.
\end{itemize}
The simultaneous decoding process continues until  no more source tokens are added in the outer loop. We define the last outer step as the terminal outer step $S$, and other outer steps as non-terminal outer steps. 

For the outer loop, we make no assumption about the value of $c_s$, while all previous work assumes $c_s=1$. This setting is more realistic because (\romannumeral1) increasing $c_s$ can reduce the number of outer steps, thus reducing computation cost; (\romannumeral2) in a real speech translation application, an ASR system may generate multiple tokens at a time. 

For the inner loop, we adapt a pretrained vanilla CNMT model to perform partial translation with two important concerns:
\begin{itemize}
    \item[1.] Prefix translation: given a source prefix $\mathbf{x}^s = \{x_1, ..., x_{\eta \left[ s\right]}\}$ and a target prefix $\mathbf{y}^s_{\tau \left[ s-1\right]} = \{y_1, ..., y_{\tau \left[ s-1\right]}\}$, how to predict the remaining target tokens? 
    \item[2.] Stopping criterion: since the NMT model is trained with full sentences, how to design the stopping criterion for it when translating partial source sentcnes?  
\end{itemize}

\subsection{Prefix Translation}\label{sec:pt}
At an outer step $s$, given encoder hidden states $\mathbf{h}^s$ for source prefix $\mathbf{x}^s= \{x_1, ..., x_{\eta \left[ s\right]}\}$ and decoder hidden states $\mathbf{z}_{\tau \left[ s-1\right]}^s$ for target prefix $\mathbf{y}_{\tau \left[ s-1\right]}^s= \{y_1, ..., y_{\tau \left[ s-1\right]}\}$, we perform prefix translation sequentially with a greedy decoder:
\begin{eqnarray}
&z_{t}^s=\phi_{\textsc{dec}}\left(y_{t-1}, z_{<t}^s, \mathbf{h}^s\right) \nonumber \\
&p(y|y_{<t},\mathbf{x}^s)\propto \exp\left[\phi_{\textsc{out}}\left(z_{t}^s\right)\right] \nonumber \\
&y_{t}=\argmax_{y} p(y|y_{<t},\mathbf{x}^s),
\end{eqnarray}
where $t$ starts from $t=\tau \left[ s-1\right]+1$. The prefix translation terminates when a stopping criterion meets, yielding a translation $\mathbf{y}^s = \{y_1, ..., y_{\tau \left[ s\right]}\}$.

However, a major problem comes from the above translation method: how can we obtain the encoder hidden states $\mathbf{h}^s$ and decoder hidden states $\mathbf{z}_{\tau \left[ s-1\right]}^s$ at the beginning of prefix translation? We propose to rebuild all encoder and decoder hidden states with 
\begin{eqnarray}
\label{eqn:enc}
\mathbf{h}^s &=& \phi_{\textsc{enc}}\big(\mathbf{x}^s\big),\\
\mathbf{z}^{s}_{\tau \left[ s-1\right]} &=& \phi_{\textsc{dec}}\big(\mathbf{y}_{\tau \left[ s-1\right]}^s,\mathbf{h}^s\big).
\end{eqnarray}
During full sentence training, all the decoder hidden states are computed conditional on the same source tokens. By rebuilding encoder and decoder hidden states, we also ensure that the decoder hidden states are computed conditional on the same source. This strategy is different from previous work that reuse previous encoder~\cite{Cho2016CanNM,Gu2017LearningTT,Dalvi2018IncrementalDA,Alinejad2018PredictionIS} or decoder~\cite{Cho2016CanNM,Gu2017LearningTT,Dalvi2018IncrementalDA,Ma2018STACLST} hidden states. We carefully compare the effect of rebuilding hidden states in Section~\ref{sec:results} and experiment results show that rebuilding all hidden states benefits translation. 

\subsection{Stopping Criterion} \label{sec:sc}
\begin{figure} 
	\centering\includegraphics[width=0.48\textwidth]{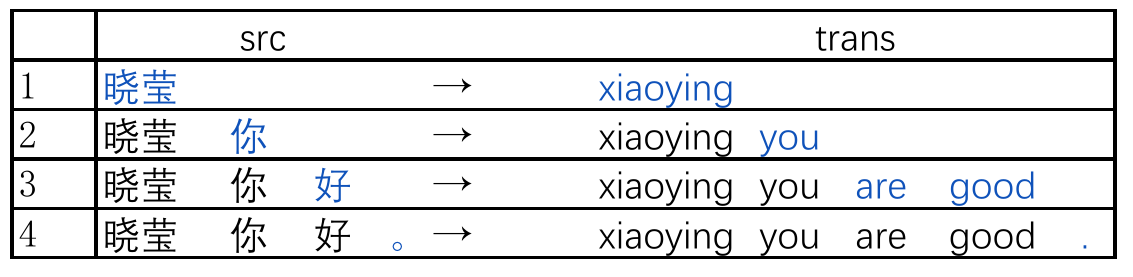}
	\caption{Failure case when using EOS alone as the stopping criterion. }\label{fig:exp_eos}
\end{figure}

In consecutive NMT, the decoding algorithm such as greedy decoding or beam search terminates when the translator predicts an EOS token or the length of the translation meets a predefined threshold (e.g. $200$). The decoding for most source sentences terminates when the translator predicts the EOS token.\footnote{We conduct greedy decoding on the validation set of WMT15 EN$\rightarrow$DE translation with fairseq-py, and find that 100\% translation terminates with EOS predicted.} In simultaneous decoding, since we use a NMT model pretrained on full sentences to translate partial source sentences, it tends to predict EOS when the source context has been fully translated. However, such strategy could be too aggressive for simultaneous translation. Fig.~\ref{fig:exp_eos} shows such an example. At outer step $2$, the translator predicts ``you EOS", emiting target token ``you". However, ``you" is not the expected translation for ``你" in the context of ``你好。". Therefore, we hope prefix translation at outer step $2$ can terminate without emitting any words. 

To alleviate such problems and do better simultaneous translation with pretrained CNMT model, we propose two novel stopping criteria for prefix translation.

\subsubsection{Length and EOS Control} \label{sec:lo}
In consecutive translation, the decoding process stops mainly when predicting EOS. In contrast, for prefix translation at non-terminal outer step, we stop the translation process when translation length is $d$ tokens behind source sentence length: $\tau[s]=\eta[s]-d$. Specifically, at the beginning of outer step $s$, we have source prefix $\mathbf{x}^s = \{x_1, ..., x_{\eta \left[ s\right]}\}$ and target prefix $\mathbf{y}_{\tau \left[ s-1\right]}^s = \{y_1, ..., y_{\tau \left[ s-1\right]}\}$. Prefix translation terminates at inner step $w_s$ when predicting an EOS token or satisfying: 
\begin{equation}
    w_s=\Big\{
\begin{array}{rcl}
& \max(0,\eta \left[ s\right]-\tau \left[ s-1\right]-d)           & {s < S}\\
& 200-\tau \left[ s-1\right]          & {s = S}
\end{array}
\end{equation}
where $d$ is a non-negative integer that determines the translation latency of the system. We call this stopping criterion as Length and EOS (LE) stopping controller. 

\subsubsection{Learning When to Stop}
\begin{figure}[!t]
\centering
\includegraphics[width=.45\textwidth]{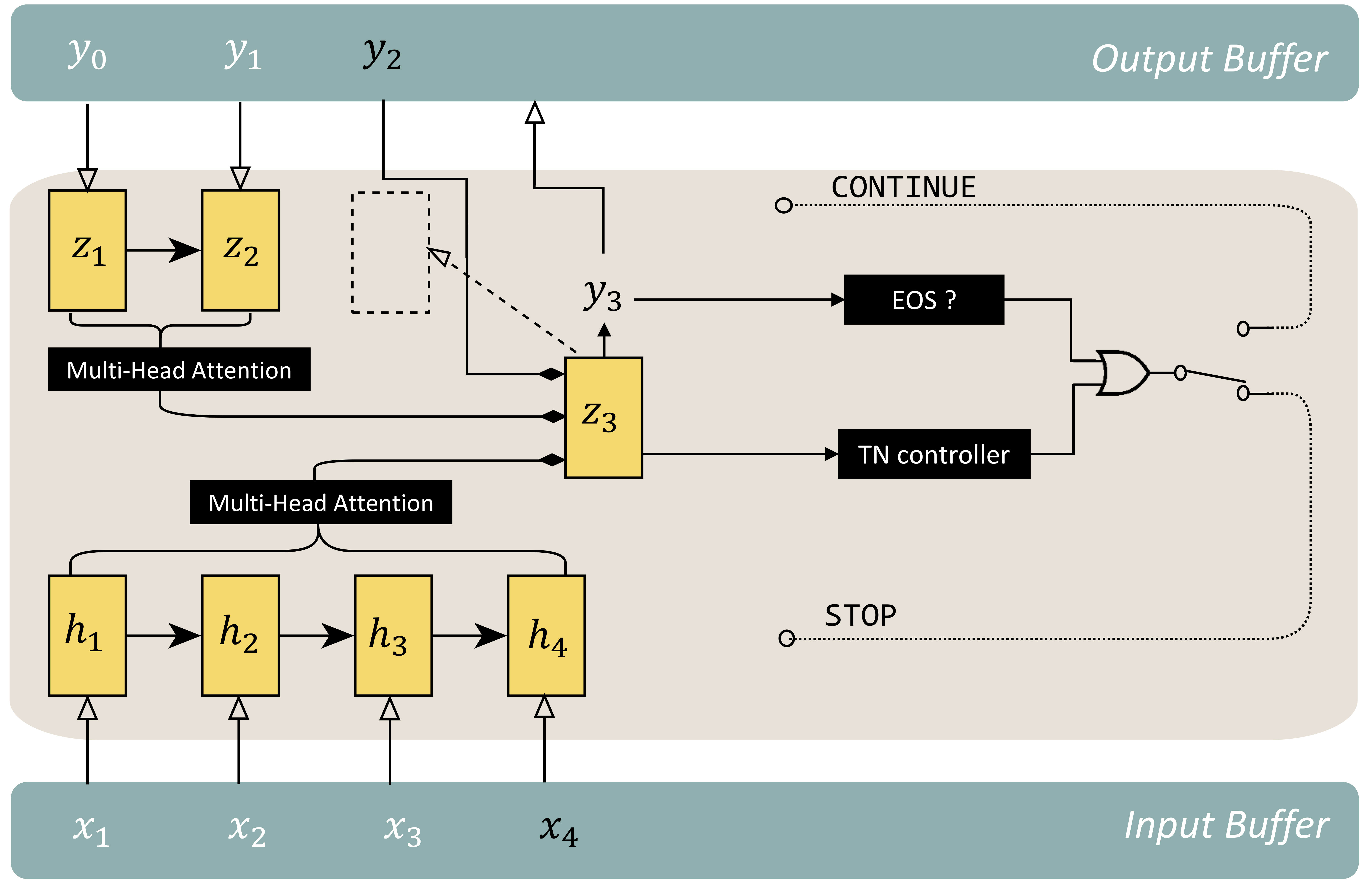}
\caption{Framework of our proposed model with the TN controller.} 
\label{fig:sys}
\end{figure}
Although simple and easy to implement, LE controller lacks the capability to learn the optimal timing with which to stop prefix translation. Therefore, we design a small trainable network called trainable (TN) stopping controller to learn when to stop prefix translation for non-terminal outer step. Fig.~\ref{fig:sys} shows the illustration. 

At each inner decoding step $k$ for non-terminal outer step $s$, the TN controller utilizes a stochastic policy $\pi_\theta$ parameterized by a neural network to make the binary decision on whether to stop translation at current step:
\begin{eqnarray}
\pi_{\theta}(a_{\tau \left[ s-1\right]+k}|z_{\tau \left[ s-1\right]+k}^s) = f_\theta(z_{\tau \left[ s-1\right]+k}^s),
\end{eqnarray}
where $z_{\tau \left[ s-1\right]+k}^s$ is the current decoder hidden state. We implement $f_\theta$ with a feedforward network with two hidden layers, followed by a softmax layer. The prefix translation stops if the TN controller predicts $a_{\tau \left[ s-1\right]+k}=1$. Our TN controller is much simpler than previous work~\cite{Gu2017LearningTT} which implements the READ/WRITE policy network using a recurrent neural network whose input is the combination of the current context vector, the current decoder state and the embedding vector of the candidate word.  

To train the TN controller, we freeze the NMT model with pretrained parameters, and optimize the TN network with policy gradient for reward maximization $\mathcal{J}= \mathbb{E}_{\pi_{\theta}}(\sum_{t=1}^{T_\tau} r_t )$. With a trained TN controller, prefix translation stops at inner decoding step $w_s$ when predicting an EOS token or satisfying: 
\begin{equation}
\Big\{
\begin{array}{rcl}
a_{\tau \left[ s-1\right]+w_s}=1     &      & {s < S}\\
w_s = 200-\tau \left[ s-1\right]  &      & {s \leq S}
\end{array}.
\end{equation}
In the following, we talk about the details of the reward function and the training with policy gradient.

\paragraph{Reward} \label{sec:delay}
To trade-off between translation quality and latency, we define the reward function at inner decoding step $k$ of outer step $s$ as:
\begin{equation}\label{eqn:reward}
    r_t = r_t^Q + \alpha \cdot r_t^D,
\end{equation}
where $t=\tau \left[ s-1\right]+k$, and $r_t^Q$ and $r_t^D$ are rewards related to quality and delay, respectively. $\alpha \geq 0$ is a hyper-parameter that we adjust to balance the trade-off between translation quality and delay.

Similar to~\citet{Gu2017LearningTT}, we utilize sentence-level BLEU~\cite{Papineni2002BleuAM,Lin2004AutomaticEO} with reward shaping~\cite{Ng1999PolicyIU} as the reward for quality:
\begin{equation}
    r_t^Q=\Big\{
\begin{array}{rcl}
 \Delta\textsc{BLEU}(\mathbf{y}^*, \mathbf{y}, t)        & {k\neq w_s\ or\ s\neq S}\\
 \textsc{BLEU}(\mathbf{y}^*, \mathbf{y})         & {k=w_s\ and\ s=S}
\end{array}
\end{equation}
where
\begin{eqnarray}
&\Delta\textsc{BLEU}(\mathbf{y}^*, \mathbf{y}, t) \qquad \qquad \qquad \qquad \nonumber \\
&= \textsc{BLEU}(\mathbf{y}^{*}, \mathbf{y}_t) - \textsc{BLEU}(\mathbf{y}^{*}, \mathbf{y}_{t-1})
\end{eqnarray} 
is the intermediate reward. Note that the higher the values of BLEU are, the more rewards the TN controller receives. Following~\citet{Ma2018STACLST}, we use average lagging (AL) as the reward for latency:
\begin{equation}
\label{eq:delay}
\begin{split}
    r_t^{D} =  \left\{\begin{array}{ll}
    0    &     k\neq w_s\ or\ s\neq S \\
    - \lfloor d(\mathbf{x}, \mathbf{y}) - d^{*} \rfloor_{+}  & k=w_s\ and\ s=S    
    \end{array} \right. 
\end{split}
\end{equation}
where 
\begin{equation}
\begin{split}
    & d\left(\mathbf{x}, \mathbf{y}\right) = \frac{1}{t_e}\sum_{t=1}^{\tau_e}{l(t)}-\frac{t-1}{\lambda}.
\end{split}
\end{equation}
$l(t)$ is the number of observed source tokens when generating the $t$-th target token, $t_e= \argmin_{t}{(l(t)=|\mathbf{x}|)}$ denotes the earliest point when the system observes the full source sentence, $\lambda=\frac{|\mathbf{y}|}{|\mathbf{x}|}$ represents the target-to-source length ratio and $d^* \geq 0$ is a hyper-parameter called target delay that indicates the desired system latency. Note that the lower the values of AL are, the more rewards the TN controller receives.

\paragraph{Policy Gradient}
We train the TN controller with policy gradient\cite{Sutton1999PolicyGM}, and the gradients are:
\begin{align}
\nabla_{\theta} \mathcal{J} = \mathbb{E}_{\pi_{\theta}}\left[\sum_{t=1}^{T_\tau} R_t\nabla_{\theta}\log \pi_{\theta}(a_t|\cdot) \right],
\end{align}
where $R_t=\sum_{i=t}^{T_\tau} r_i$ is the cumulative future rewards for the current decision. We can adopt any sampling approach \cite{chen2017teacher,chen2018zero,shen2018zero} to estimate the expected gradient. In our experiments, we randomly sample multiple action trajectories from the current policy $\pi_{\theta}$ and estimate the gradient with the collected accumulated reward. We try the variance reduction techniques by subtracting a baseline average reward estimated by a linear regression model from $R_t$ and find that it does not help to improve the performance. Therefore, we just normalize the reward in each mini-batch without using baseline reward for simplicity.

\section{Experiments}
\subsection{Settings}
\paragraph{Dataset} 
\begin{table}[!t]
	\centering
	\begin{tabular}{l | r r r}
    	\hline
		Dataset & Train & Validation & Test \\ \hline
        IWSLT16 & 193,591 & 993 & 1,305  \\ 
        WMT15 & 3,745,796 & 3,003 & 2,169 \\ 
        NIST & 1,252,977 & 878 & 4,103  \\ \hline
	\end{tabular}
	\caption{\# sentences in each dataset.}\label{table:stat}
\end{table}
We compare our approach with the baselines on WMT15 German-English\footnote{http://www.statmt.org/wmt15/} (DE-EN) translation in both directions. This is also the most widely used dataset to evaluate SNMT's performance~\cite{Cho2016CanNM,Gu2017LearningTT,Ma2018STACLST,Arivazhagan2019MonotonicIL,zheng-etal-2019-simpler}. To further evaluate our approach's efficacy in trading off translation quality and latency on other language pair and spoken language, we also conduct experiments with the proposed LE and TN methods on NIST Chinese-to-English\footnote{These sentence pairs are mainly extracted from LDC2002E18, LDC2003E07, LDC2003E14, Hansards portion of LDC2004T07, LDC2004T08 and LDC2005T06} (ZH$\rightarrow$EN) translation and IWSLT16 German-English\footnote{https://workshop2016.iwslt.org/} (DE-EN) translation in both directions. For WMT15, we use newstest2014 for validation and newstest2015 for test. For NIST, we use MT02 for validation, and MT05, MT06, MT08 for test. For IWSLT16, we use tst13 for validation and tst14 for test. All the data is tokenized and segmented into subword symbols using byte-pair encoding \cite{Sennrich2016NeuralMT} to restrict the size of the vocabulary. We use 40,000 joint merge operations on WMT15, and 24,000 on IWSLT16. For NIST, we use 30,000 merge operations for source and target side separately. Without explicitly mention, we simulate simultaneous translation scenario at inference time with these datasets by assuming that the system observes one new source token at each outer step, i.e., $c_s=1$. Table~\ref{table:stat} shows the data statistics.

\paragraph{Pretrained NMT Model}
\begin{figure*}[!t]
\centering
\includegraphics[width=0.95\textwidth]{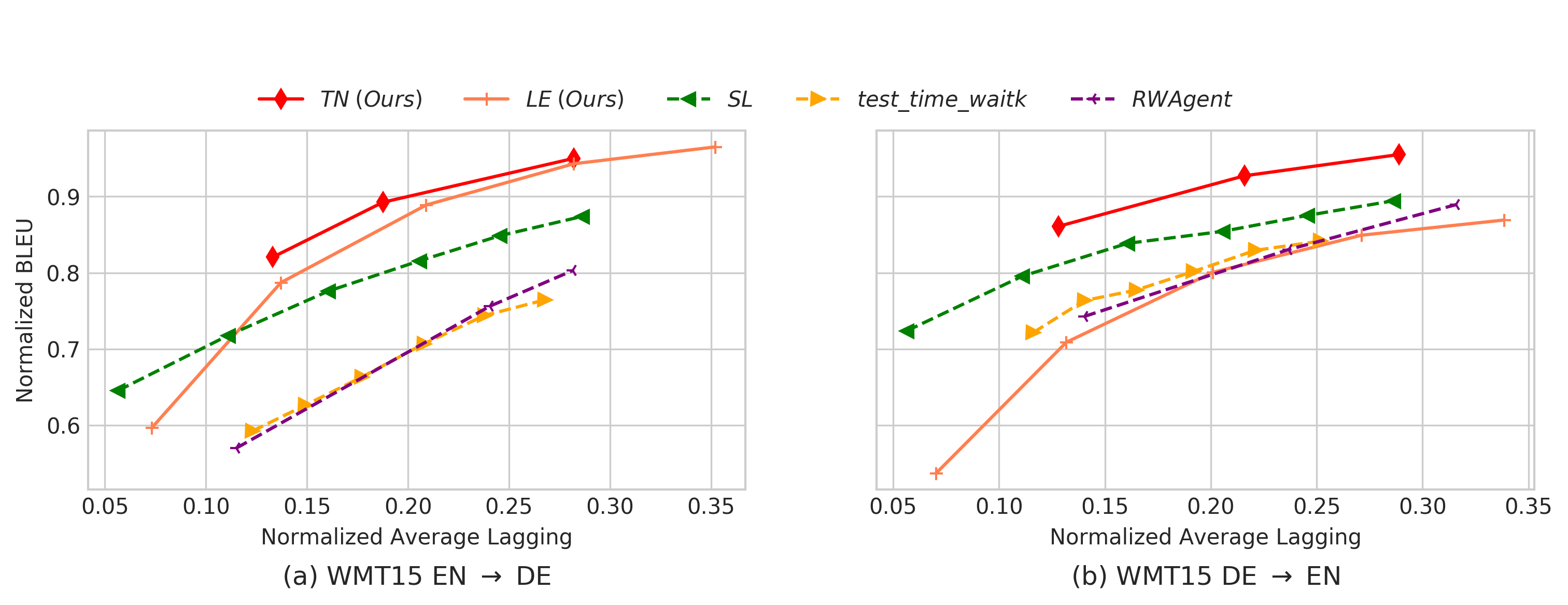}
\caption{Comparison with the baselines on the test set of WMT15 EN$\rightarrow$DE and WMT15 DE$\rightarrow$EN translations. The shown points from left to right on the same line are the results of simultaneous greedy decoding with $d^* \in \{2,5,8\}$ for TN, $d \in \{0,2,4,6,8\}$ for LE, $\rho \in \{0.65,0.6,0.55,0.5,0.45,0.4\}$ for SL, $k \in \{1,3,5,7,9\}$ for test\_time\_waitk and $CW \in \{2,5,8\}$ for RWAgent. The scores of \textbf{Greedy} decoding: BLEU=$25.16$, AL=$28.10$ for WMT15 EN$\rightarrow$DE translation and BLEU=$26.17$, AL=$31.20$ for WMT15 DE$\rightarrow$EN translation.}

\label{fig:wmt_cpr}
\end{figure*}

\begin{figure*}[!t]
\centering
\includegraphics[width=0.95\textwidth]{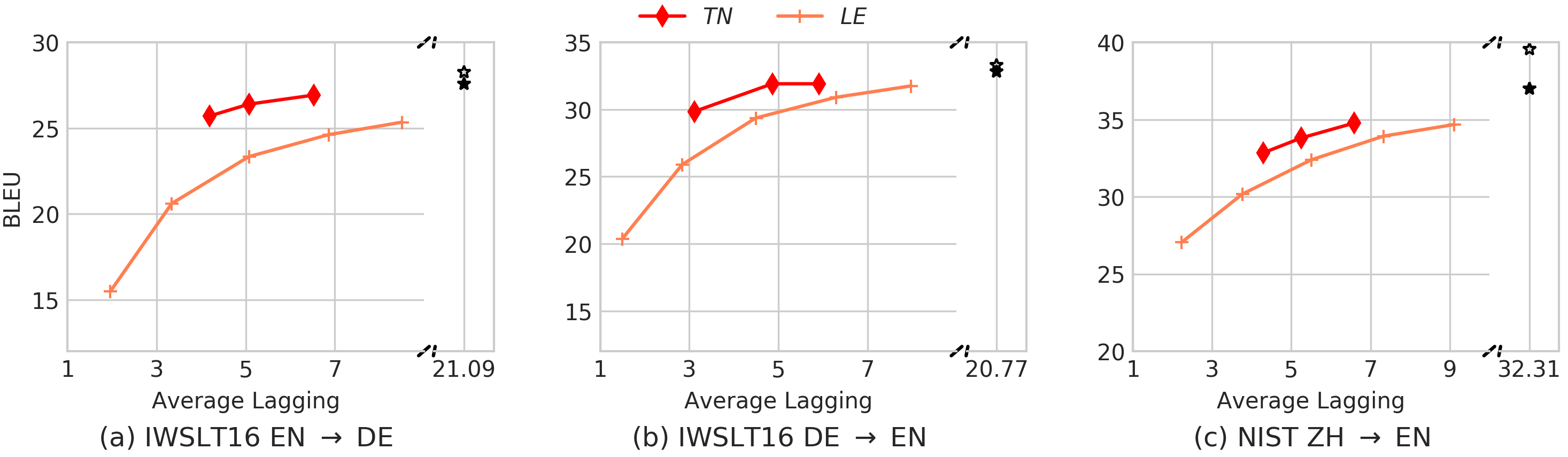}
\caption{Performance on the test set of IWSLT16 EN$\rightarrow$DE translation, IWSLT16 DE$\rightarrow$EN translation and NIST ZH$\rightarrow$EN translation. The shown points from left to right on the same line are the results of $d^* \in \{2,5,8\}$ for TN and $d \in \{ 0,2,4,6,7\}$ for LE. \ding{72}\ding{73}:full-sentence (greedy and beam-search).} 
\label{fig:iwslt_nist}
\end{figure*}

\begin{figure}[!t]
\centering
\includegraphics[width=0.45\textwidth]{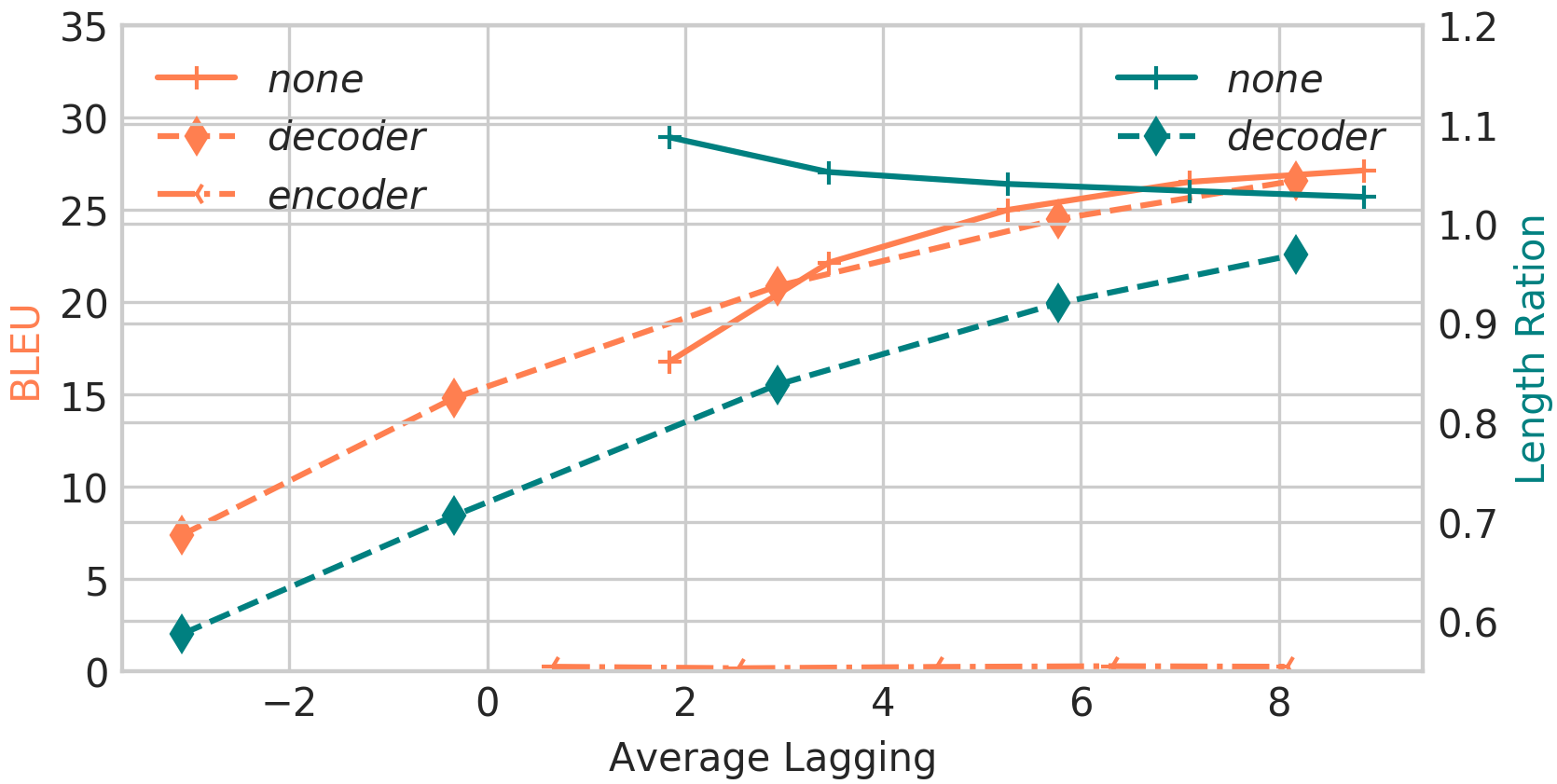}
\caption{Comparison of whether to reuse previous encoder or decoder hidden states on WMT15 EN$\rightarrow$DE test set with the LE controller. The left Y axis is the BLEU score and the right Y axis is the length ratio: the translation length divided by the reference length. The points on the same line are the results of $d \in \{0,2,4,6,8\}$. \textit{none}: rebuild all encoder/decoder hidden states; \textit{decoder}: reuse decoder hidden states and rebuild all encoder hidden states; \textit{encoder}: reuse previous encoder hidden states and rebuild all decoder hidden states.}
\label{fig:wmt_ende}
\end{figure}

\begin{figure*}[!t]
\centering
\includegraphics[width=0.95\textwidth]{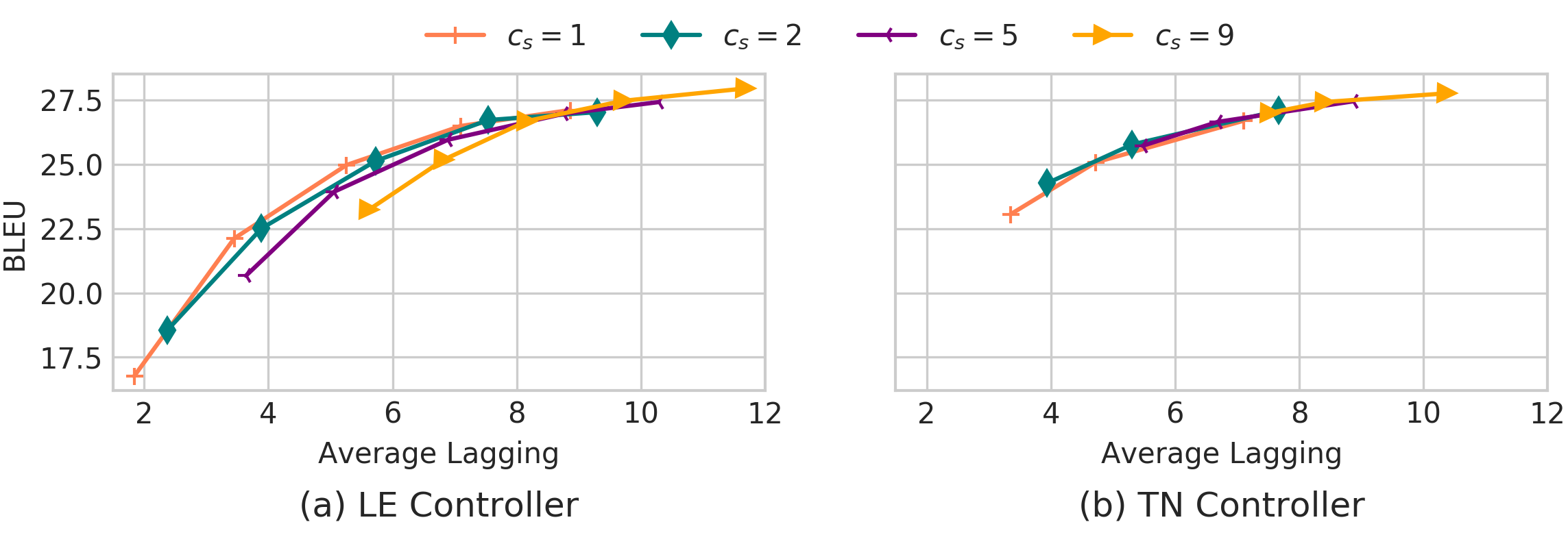}
\caption{Performance on the test set of WMT15 EN$\rightarrow$DE translation with different input buffer update schedule. Points on the same line are obtained by increasing $d \in {0,2,4,6,8}$ for (a) and $d^* \in {2,5,8}$ for (b).} 
\label{fig:wmt_ende_ns}
\end{figure*}

\begin{figure*}[!t]
    \centering
    \includegraphics[width=0.95\textwidth]{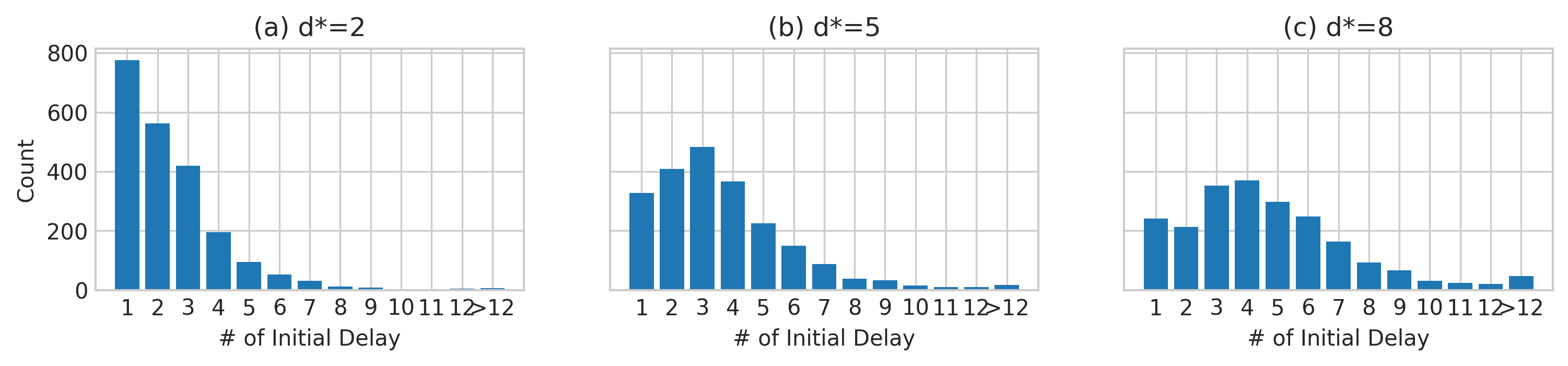}
    \caption{Number of observed source tokens before emitting the first target token for the TN controller on the test set of WMT15 EN$\rightarrow$DE translation.} 
    \label{fig:count}
\end{figure*}
We use Transformer~\cite{Vaswani2017AttentionIA} trained with maximum likelihood estimation as the pretrained CNMT model and implement our method based on fairseq-py.\footnote{https://github.com/pytorch/fairseq} We follow the setting in \texttt{transformer\_iwslt\_de\_en} for IWSLT16 dataset, and \texttt{transformer\_wmt\_en\_de} for WMT15 and NIST dataset. Fairseq-py adds an EOS token for all source sentences during training and inference. Therefore, to be consistent with the CNMT model implemented with fairseq-py, we also add an EOS token at the end of the source prefix for prefix translation and find that the EOS helps translation.

\paragraph{TN Controller} To train the TN controller, we use a mini-batch size of 8,16,16 and sample 5,10,10 trajectories for each sentence pair in a batch for IWSLT16, WMT15 and NIST, respectively. We set the number of newly observed source tokens at each outer step to be $1$ during the training for simplicity. We set $\alpha$ to be $0.04$, and $d^*$ to be $2,5,8$. All our TN controllers are trained with policy gradient using Adam optimizer~\cite{Kingma2015AdamAM} with 30,000 updates. We select the last model as our final TN controller.

\paragraph{Baseline} 
We compare our model against three baselines that utilize a pretrained CNMT model to perform simultaneous translation:
\begin{itemize}
    \item \textbf{test\_time\_waitk}~\cite{Ma2018STACLST}: the method that decodes with a waitk policy with a CNMT model. We report the results when $k\in \{1,3,5,7,9\}$.
    \item \textbf{SL}~\cite{zheng-etal-2019-simpler}: the method that adapts CNMT to SNMNT by learning an adaptive READ/WRITE policy from oracle READ/WRITE sequences generated with heuristics. We report the results with threshold $\rho \in \{0.65,0.6,0.55,0.5,0.45,0.4\}$.
    \item \textbf{RWAgent}~\cite{Gu2017LearningTT}: the adaptation of \citet{Gu2017LearningTT}'s  full-sentence model and reinforced READ/WRITE policy network to Transformer by~\citet{Ma2018STACLST}. We report the results when using $CW \in \{2,5,8\}$ as the target delay.
\end{itemize}

We report the result with $d\in \{0,2,4,6,8\}$ for our proposed LE method and $d^*\in \{2,5,8\}$ for our proposed TN method. For all baselines, we cite the results reported in~\citet{zheng-etal-2019-simpler}. \footnote{Since \citet{zheng-etal-2019-simpler} did not mention the details of data preprocessing, we cannot compare the BLEU and AL scores directly with theirs. Therefore, we normalize the BLEU and AL scores with its corresponding upper bound, i.e. the BLEU and AL scores obtained when the pretrained Transformer performs standard greedy decoding (\textbf{Greedy}).}

\subsection{Results}\label{sec:results}
We compare our methods with the baselines on the test set of WMT15 EN$\rightarrow$DE and DE$\rightarrow$EN translation tasks, as shown in Fig.~\ref{fig:wmt_cpr}. The points closer to the upper left corner indicate better overall performance, namely low latency and high quality. We observe that as latency increases, all methods improve in quality. the TN method significantly outperforms all the baselines in both translation tasks, demonstrating that it indeed learns the appropriate timing to stop prefix translation. LE outperforms the baselines on WMT15 EN$\rightarrow$DE translation at high latency region and performs similarly or worse on other cases. 

We show the methods' efficacy in trading off quality and latency on other language pair and spoken language in Fig.~\ref{fig:iwslt_nist}. TN outperforms LE on all translation tasks, especially at the low latency region. It obtains promising translation quality with acceptable latency: with a lag of $<7$ tokens, TN obtains 96.95\%, 97.20\% and 94.03\% BLEU with respect to consecutive greedy decoding for IWSLT16 EN$\rightarrow$DE, IWSLT16 DE$\rightarrow$EN and NIST ZH$\rightarrow$EN translations, respectively.

\subsection{Analyze}
\begin{figure}[!t]
\centering
\includegraphics[width=0.45\textwidth]{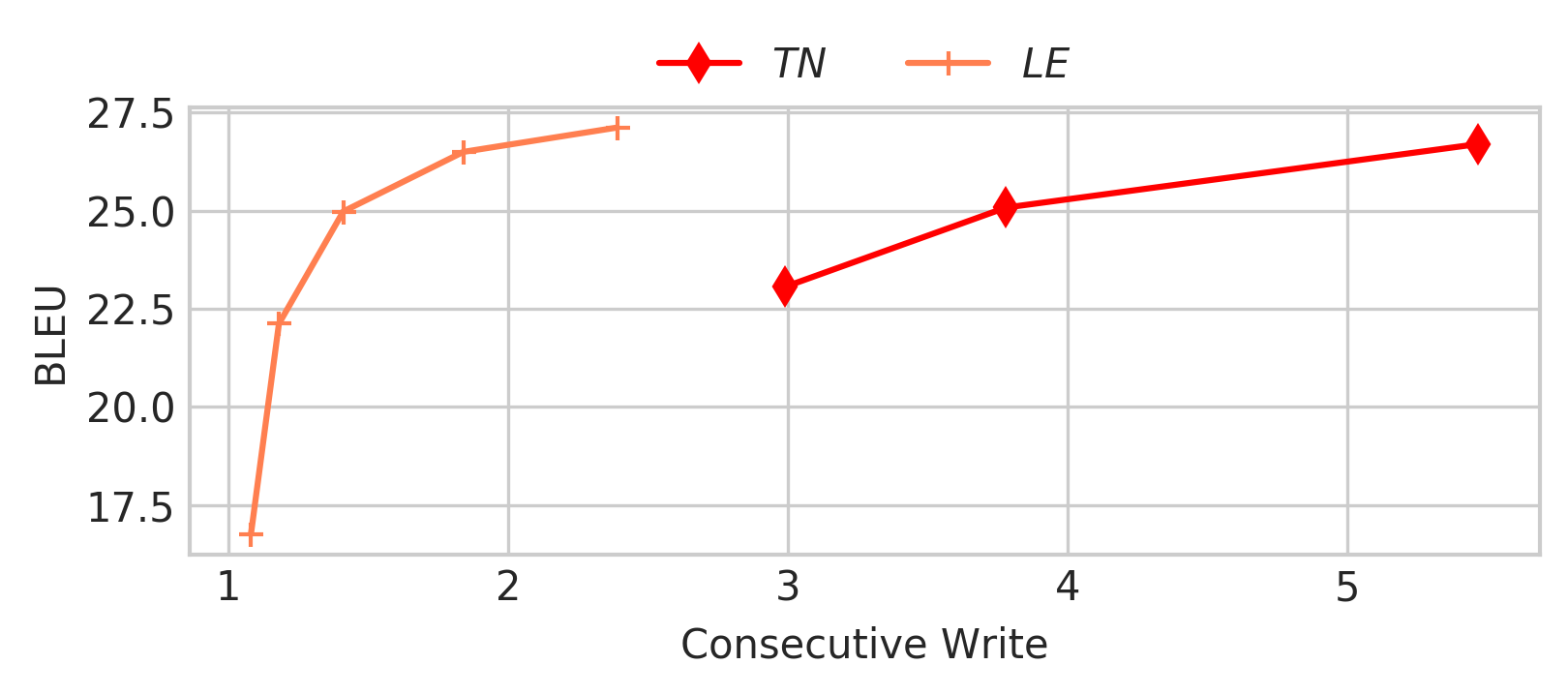}
\caption{Average consecutive write length on the test set of WMT15 EN$\rightarrow$DE translation.} 
\label{fig:cw}
\end{figure}

\begin{figure*}[!htb]
\centering
\resizebox{1.0\textwidth}{!}{%
\setlength{\tabcolsep}{.7pt}
\begin{tabu}{ c | l l l l l l l l l  }
\rowfont{\small}
\hline
  & 1 & 2 & 3 & 4 & 5 & 6 & 7 & 8 & 9     \\ \hline \hline
         & 吴邦国 & 出席 & 签字 & 仪式 & 并 & 在 &  协议 & 上 & 签字   \\
  \hline
  LE  &  &  & &  & wu & bangguo & attended & the & signing  ceremony and signed the agreement \\
  TN  & &  & & & wu bangguo attended the signing ceremony &  & &  & and signed the agreement
  \\
  Greedy  & &  &  & &  & & & &  wu bangguo attended the signing ceremony and signed the agreement
 \\
 Ref  & &  &  & &  & & & &  wu bangguo attends signing ceremony and signs agreement \\
 \hline \hline
         & NATO & does & not & want & to & break &  agreements & with & Russia   \\
  \hline
  LE  &  &  & &  & Die & NATO & möchte & keine & Abkommen mit Russland brechen \\
  TN  & & Die NATO & & & will &  & & keine  Abkommen & mit Russland brechen
  \\
  Greedy  & &  &  & &  & & & &  Die NATO möchte keine Abkommen mit Russland brechen
 \\
 Ref  & &  &  & &  & & & & NATO will Vereinbarungen mit Russland nicht brechen
 \\
 \hline
\end{tabu}
}
\caption{Translation examples from the test set of NIST ZH$\rightarrow$EN (example 1) and WMT15 EN$\rightarrow$DE translation (example 2). We compare LE with $d=4$ and TN with $d^*=5$ because these two models achieve similar latency. Greedy and Ref represent the greedy decoding result from consecutive translation and the reference, respectively.}
\label{fig:exp}
\end{figure*}
We analyze the effect of different ways to obtain the encoder and decoder hidden states at the beginning of prefix translation with the LE controller. Fig.~\ref{fig:wmt_ende} shows the result. We try three variants: a) dynamically rebuild all encoder/decoder hidden states (\textit{none}); b) reuse decoder hidden states and rebuild all encoder hidden states (\textit{decoder}); c) reuse previous encoder hidden states and rebuild all decoder hidden states (\textit{encoder}). The left Y axis and X axis show BLEU-vs-AL curve. We observe that if reusing previous encoder hidden states (\textit{encoder}), the translation fails. We ascribe this to the discrepancy
between training and decoding for the encoder. We also observe that when $d\in 0,2$, reusing decoder hidden states (\textit{decoder}) obtain negative AL. To analyze this, we plot the translation to reference length ratio versus AL curve with the right Y axis and X axis. It shows that with \textit{decoder}, the decoding process stops too early and generates too short translations. Therefore, to avoid such problem and to be consistent with the training process of the CNMT model, it is important to dynamically rebuild all encoder/decoder hidden states for prefix translation.

Since we make no assumption about the $c_s$, i.e., the number of newly observed source tokens at each outer step, we also test the effect of different $c_s$. Fig.~\ref{fig:wmt_ende_ns} shows the result with the LE and TN controllers on the test set of WMT15 EN$\rightarrow$DE translation. We observe that as $c_s$ increases, both LE and TN trend to improve in quality and worsen in latency. When $c_s=1$, LE controller obtains the best balance between quality and latency. In contrast, TN controller obtains similar quality and latency balance with different $c_s$, demonstrating that TN controller successfully learns the right timing to stop regardless of the input update schedule. 

We also analyze the TN controller's adaptability by monitoring the initial delay, i.e., the number of observed source tokens before emitting the first target token, on the test set of WMT15 EN$\rightarrow$DE translation, as shown in Fig.~\ref{fig:count}. $d^*$ is the target delay measured with AL (used in Eq.~\ref{eq:delay}). It demonstrates that the TN controller has a lot of variance in it's initial delay. The distribution of initial delay changes with different target delay: with higher target delay, the average initial delay is larger. For most sentences, the initial delay is within $1-7$.

In speech translation, listeners are also concerned with long silences during which no translation occurs. Following~\citet{Gu2017LearningTT,Ma2018STACLST}, we use Consecutive Wait (CW) to measure this:
\begin{equation}
CW(\mathbf{x},\mathbf{y}) = \frac{\sum_{s=1}^{S} c_s}{\sum_{s=1}^S \mathbbm{1}_{w_s>0}}.
\end{equation}
Fig.~\ref{fig:cw} shows the BLEU-vs-CW plots for our proposed two methods. The TN controller has higher CW than the LE controller. This is because TN controller prefers consecutive updating output buffer (e.g., it often produces $w_s$ as $0\ 0\ 0\ 0\ 3\ 0\ 0\ 0\ 0\ 0\ 5\ 0\ 0\ 0\ 0\ 4\ ...$) while the LE controller often updates its output buffer following the input buffer (e.g., it often produces $w_s$ as $0\ 0\ 0\ 0\ 1\ 1\ 1\ 1\ 1\ 1\ ...$ when $d=4$). Although larger than LE, the CW for TN ($< 6$) is acceptable for most speech translation scenarios.  

\subsection{Translation Examples} 
Fig.~\ref{fig:exp} shows two translation examples with the LE and TN controllers on the test set of NIST ZH$\rightarrow$EN and WMT15 EN$\rightarrow$DE translation. In manual inspection of these examples and others, we find that the TN controller learns a conservative timing for stopping prefix translation. For example, in example $1$, TN outputs translation \textit{``wu bangguo attended the signing ceremony''} when observing \textit{``吴邦国\ 出席\ 签字\ 仪式\ 并''}, instead of a more radical translation \textit{``wu bangguo attended the signing ceremony and''}. Such strategy helps to alleviate the problem of premature translation, i.e., translating before observing enough future context.

\section{Related Work}
A number of works in simultaneous translation divide the translation process into two stages. A segmentation component first divides the incoming text into segments, and then each segment is translated by a translator independently or with previous context. The segmentation boundaries can be predicted by prosodic pauses detected in speech \cite{Fgen2007SimultaneousTO,Bangalore2012RealtimeIS}, linguistic cues \cite{Sridhar2013SegmentationSF,Matusov2007ImprovingST}, or a classifier based on alignment information \cite{Siahbani2014IncrementalTU,Yarmohammadi2013IncrementalSA} and translation accuracy \cite{Oda2014OptimizingSS,Grissom2014DontUT,Siahbani2018SimultaneousTU}. 

Some authors have recently endeavored to perform simultaneous translation in the context of NMT. \citet{niehues2018low,arivazhagan2020re} adopt a re-translation approach where the source is repeatedly translated from scratch as it grows and propose methods to improve translation stability. \citet{Cho2016CanNM}; \citet{Dalvi2018IncrementalDA}; \citet{Ma2018STACLST} introduce a manually designed criterion to control when to translate. \citet{Satija2016}; \citet{Gu2017LearningTT}; \citet{Alinejad2018PredictionIS} extend the criterion into a trainable agent in a reinforcement learning framework. However, these work either develop sophisticated training frameworks explicitly designed for simultaneous translation~\cite{Ma2018STACLST} or fail to use a pretrained consecutive NMT model in an optimal way~\cite{Cho2016CanNM,Dalvi2018IncrementalDA,Satija2016,Gu2017LearningTT,Alinejad2018PredictionIS,zheng-etal-2019-simpler}. In contrast, our work is significantly different from theirs in the way of using pretrained consecutive NMT model to perform simultaneous translation and the design of the two stopping criteria.

\section{Conclusion}
We have presented a novel framework for improving simultaneous translation with a pretrained consecutive NMT model. The basic idea is to translate partial source sentence with the consecutive NMT model and stops the translation with two novel stopping criteria. Extensive experiments demonstrate that our method with trainable stopping controller outperforms the state-of-the-art baselines in balancing between translation quality and latency. 

\section*{Acknowledgments}
We thank the anonymous reviewers for their insightful feedback on this work. Yun Chen is partially supported by the Fundamental Research Funds for the Central Universities and the funds of Beijing Advanced Innovation Center for Language Resources (No. TYZ19005).


\bibliographystyle{acl_natbib}
\bibliography{aacl-ijcnlp2020}
\end{CJK*}
\end{document}